\documentclass[sigconf]{acmart}
\acmSubmissionID{1234}

\usepackage{booktabs} 

\usepackage[ruled]{algorithm2e} 

\SetAlFnt{\small}
\SetAlCapFnt{\small}
\SetAlCapNameFnt{\small}
\SetAlCapHSkip{0pt}

\copyrightyear{2026}
\acmYear{2026}
\setcopyright{cc}
\setcctype{by}
\acmConference[SA Technical Communications '26]{SIGGRAPH Asia 2026 Technical Communications}{December 01--04, 2026}{Kuala Lumpur, Malaysia}
\acmBooktitle{SIGGRAPH Asia 2026 Technical Communications (SA Technical Communications '26), December 01--04, 2026, Kuala Lumpur, Malaysia}
\acmDOI{10.1145/3829339.3847859}
\acmISBN{979-8-4007-2841-9/2026/12}

\usepackage{booktabs}
\usepackage{xcolor}
\usepackage{colortbl}
\usepackage{graphicx}

\definecolor{baselinecolor}{RGB}{255,220,185}
\definecolor{ourscolor}{RGB}{255,180,180}

\begin{document}
\def\titlePrefix{4DGS-Fixer}
\title[4DGS-Fixer]{\titlePrefix{}: Generative Sparse-View 4D Gaussian Splatting with Iterative Refinement Guided by Video Diffusion Priors}

\author{Haitao Huang}\authornote{Joint first author.}
\email{marin.huanght@goertek.com}
\affiliation{
  \institution{Goertek Alpha Labs}
  \city{Shanghai}
  \country{China}
}

\author{Shenghao Zhao}\authornotemark[1]
\email{Shenghao.Zhao@singaporetech.edu.sg}
\affiliation{
  \institution{Singapore Institute of Technology}
  \city{Singapore}
  \country{Singapore}
}

\author{Boyuan Tian}
\email{talon.tian@goertek.com}
\affiliation{
  \institution{Goertek Alpha Labs}
  \city{Nanjing}
  \country{China}
}

\author{Shin-Fang Chng}
\email{shinfang.chng@goertekusa.com}
\affiliation{
  \institution{Goertek Alpha Labs}
  \city{Santa Clara}
  \country{USA}
}

\author{Songlin Yang}
\email{syangds@connect.ust.hk}
\orcid{0000-0003-3403-376X}
\affiliation{
  \institution{The Hong Kong University of Science and Technology}
  \city{Hong Kong}
  \country{Hong Kong SAR, China}
}

\author{Sheila Lim Yann Tsern}
\email{sheilalimyanntsern@gmail.com}
\affiliation{
  \institution{Singapore Institute of Technology}
  \city{Singapore}
  \country{Singapore}
}

\author{Huangying Zhan}\authornote{Corresponding author.}
\email{zhanhuangying.work@gmail.com}
\affiliation{
  \institution{Goertek Alpha Labs}
  \city{Santa Clara}
  \country{USA}
}

\author{Yi Xu}
\email{yi.xu.purdue@gmail.com}
\affiliation{
  \institution{Goertek Alpha Labs}
  \city{Santa Clara}
  \country{USA}
}

\author{Anyi Rao}
\email{anyirao@ust.hk}
\affiliation{
  \institution{The Hong Kong University of Science and Technology}
  \city{Hong Kong}
  \country{Hong Kong SAR, China}
}

\author{Frank Guan}
\email{frank.guan@singaporetech.edu.sg}
\affiliation{
  \institution{Singapore Institute of Technology}
  \city{Singapore}
  \country{Singapore}
}

\renewcommand\shortauthors{Huang, et al.}

\begin{abstract}
This paper addresses dynamic scene synthesis from sparse-view videos. Existing methods employ geometric priors, adaptive optimization, or density-control strategies to improve 4D Gaussian modeling under sparse observations. However, they cannot fundamentally resolve the ill-posed problem caused by insufficient observations and missing scene information. 
Moreover, sparse-view 4D Gaussian Splatting (4DGS) often suffers from poor geometric initialization: with only a few input views, COLMAP typically reconstructs sparse and incomplete point clouds, leaving large scene regions without sufficient Gaussian support and making them difficult to recover through subsequent optimization. To address these limitations, we propose a novel iterative refinement framework based on a video diffusion model to improve the completeness and consistency of dynamic 4D scenes. Specifically, we first estimate multi-view depth maps and fuse them into dense point clouds to provide more complete geometric initialization for a dynamic 4DGS representation. We then employ a pretrained video restoration model to refine sequences rendered along novel camera trajectories at different time steps. The restored sequences serve as pseudo-supervision to regularize and iteratively refine the 4DGS representation. Experiments on a widely used benchmark dataset show that our method substantially outperforms existing baselines, improving PSNR by nearly 2dB over the previous best method.
\end{abstract}

%
%


%
%


\maketitle

\section{Introduction}
\label{sec:intro}

Dynamic novel-view synthesis reconstructs time-varying 3D scenes from multi-view videos and renders them from unseen viewpoints. Recent 
4D Gaussian Splatting (4DGS) methods~\cite{li2024spacetimegaussianfeaturesplatting,yang2024realtimephotorealisticdynamicscene} enable efficient and high-quality dynamic view rendering, but typically rely on densely distributed and synchronized cameras. 
With only a few input views -- as is common in practical capture settings -- the problem becomes severely ill-posed because large portions of the scene are weakly observed or entirely missing.

Existing sparse-view 4DGS typically employ geometric priors, adaptive optimization and density-controlled strategies to improve reconstruction. However, they remain constrained by the visual evidence available in the input views and therefore cannot recover missing scene information. 
Generative priors offer a promising way to infer plausible information beyond the observed views. However, most existing generative reconstruction methods~\cite{Kong_2025_CVPR}  focus on static scenes, while dynamic approaches~\cite{jin2025diffuman4d4dconsistenthuman} are often restricted to specific domains such as humans.

Moreover, sparse 4DGS also suffers from poor initialization. Limited view overlap, non-rigid motion, self-occlusion and low-texture make feature matching and triangulation unreliable, causing COLMAP to produce sparse and incomplete point clouds that biased towards highly texture regions. As Gaussian densification mainly expands around existing primitives, regions missing from the initialization are difficult to recover during optimization.


\begin{figure*}[t]
    \centering
    \includegraphics[width=\linewidth]{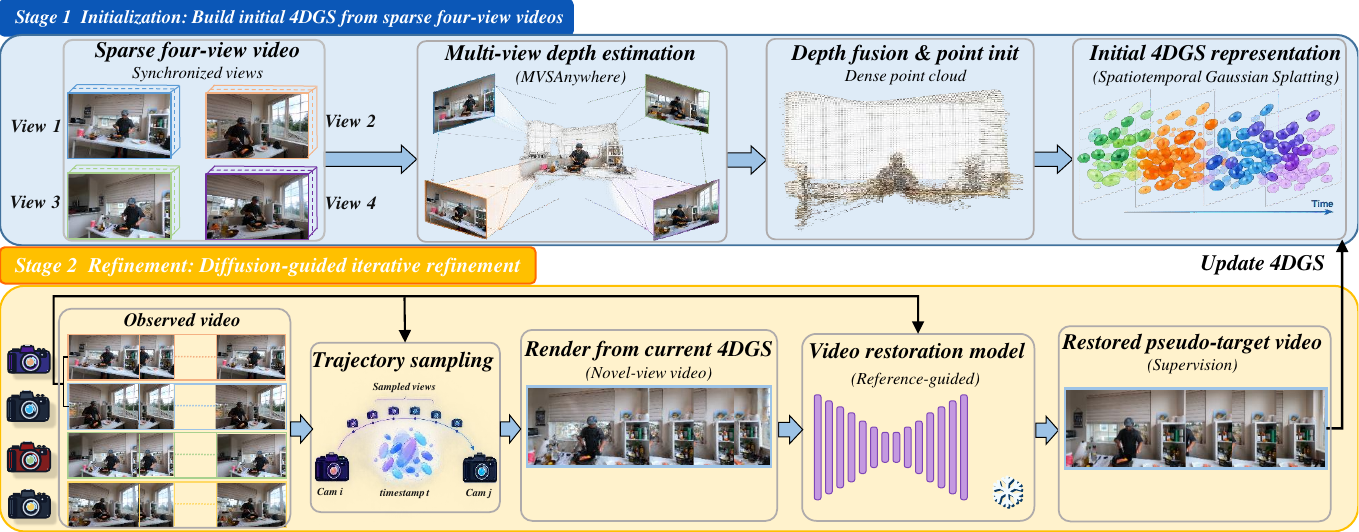}

    \caption{\textbf{Overview of 4DGS-Fixer.}
    Dense multi-view point clouds initialize the 4DGS representation,
    while restored novel-trajectory renderings provide cached pseudo-supervision for iterative refinement.}

    \label{fig:pipeline}
\end{figure*}



We propose \textbf{\titlePrefix{}}, a two-stage framework combining dense geometric initialization with video-diffusion-guided refinement.
First, we fuse depth maps estimated by multi-view stereo (MVS) into dense point clouds, providing more complete geometric coverage than COLMAP.
We initialize 4DGS with these point clouds and optimize it on the observed views.
We then render videos along novel trajectories between observed camera pairs and restore them using a pretrained video restoration model, with observed views anchoring cross-view consistency.
The restored videos provide pseudo-supervision for iterative 4DGS refinement (Fig.~\ref{fig:pipeline}).

We summarize our contributions as follows:
\begin{itemize}
    \item We propose \titlePrefix{}, a generative 4DGS framework for general sparse-view dynamic scene reconstruction.
    
    \item We introduce a dense initialization strategy that fuses multi-view depth predictions into dense point clouds, improving both reconstruction quality and initialization efficiency. 

    \item We develop a diffusion-guided refinement strategy that restores video rendered along sampled novel-view trajectories using a reference-guided video restoration model and uses them as pseudo-supervision to iteratively improve the spacetime Gaussian representation.

    \item We demonstrate that \titlePrefix{} achieves state-of-the-art performance on the Neural3DV benchmark, outperforming recent state-of-the-art, 4C4D by 1.89 dB in PSNR. 
    
\end{itemize}

\section{Related Work}
\label{sec:rel-work}

Dynamic NeRFs model temporal variations using deformation fields or factorized
space--time representations~\cite{fridovichkeil2023kplanes}, while 4DGS
methods achieve efficient dynamic reconstruction with spatiotemporal
Gaussians~\cite{li2024spacetimegaussianfeaturesplatting}.
These methods typically require dense and synchronized
camera views. Sparse-view methods alleviate this requirement through neural
or geometric regularization~\cite{zhou20264c4d4camera4d}, but remain constrained by
the limited observations in the input views.

Generative priors have recently been used to recover missing observations by synthesizing novel views with image or video diffusion models and distilling them into the 3D representation. However, most existing methods focus on static scenes~\cite{Kong_2025_CVPR,yin2025gsfixerimproving3dgaussian}. 
While Diffuman4D~\cite{jin2025diffuman4d4dconsistenthuman} tackles dynamic reconstruction, it is restricted to human-centric scenes. 
\section{Method}
\label{sec:method}

We propose \titlePrefix{}, a novel generative sparse 4DGS framework for high-fidelity dynamic scene reconstruction from only four input views. 
Sec.~\ref{sec:init} first introduces dense point-cloud initialization using a multi-view depth estimation network. 
Sec.~\ref{sec:diffusion} then presents our iterative refinement approach based on a video restoration model.

\subsection{Dense geometric initialization}
\label{sec:init}

The original spacetime Gaussian Splatting (STGS) pipeline initializes the Gaussian representation using the
sparse point cloud reconstructed by COLMAP. In our pipeline, COLMAP is used only at the first timestamp to estimate the camera intrinsics and extrinsics.
Since the four cameras remain fixed throughout the sequence, the estimated camera parameters are shared across all timestamps. 
At each timestamp \(t\), the four synchronized frames and their corresponding camera parameters are passed to
MVSAnywhere~\cite{izquierdo2025mvsanywherezeroshotmultiviewstereo}, which predicts one depth map for each view. 
These depth maps are then fused into a dense colored point cloud to provide frame-wise geometric initialization for 4DGS optimization. 
Compared with sparse COLMAP point clouds, this dense initialization offers more complete geometric coverage and enables more efficient optimization.

Specifically, let \(D_{i,t}(\mathbf{u})\) denote the depth predicted for pixel
\(\mathbf{u}=(u,v)\) in view \(i\) at timestamp \(t\), and let
\(\mathbf{K}_i\) and \(\mathbf{T}_{w \leftarrow c_i}\) denote the camera
intrinsic matrix and the camera-to-world transformation, respectively. Each
valid depth pixel is back-projected into the world coordinate system as
\begin{equation}
\mathbf{x}_{i,t}(\mathbf{u})
=
\mathbf{T}_{w \leftarrow c_i}
\begin{bmatrix}
D_{i,t}(\mathbf{u})\mathbf{K}_i^{-1}\tilde{\mathbf{u}} \\
1
\end{bmatrix},
\qquad
\tilde{\mathbf{u}}=
\begin{bmatrix}
u & v & 1
\end{bmatrix}^{\mathsf{T}}.
\label{eq:depth_backprojection}
\end{equation}

The dense point cloud at time $t$ is obtained by merging the back-projected points from all four views:
%
\begin{equation}
\mathcal{P}_{t}
=
\mathcal{F}
\left(
\bigcup_{i=1}^{4}
\left\{
\left(
\mathbf{x}_{i,t}(\mathbf{u}),
\mathbf{I}_{i,t}(\mathbf{u})
\right)
\;\middle|\;
\mathbf{u}\in\Omega_i,\,
M_{i,t}(\mathbf{u})=1
\right\}
\right),
\label{eq:depth_fusion}
\end{equation}

where \(\mathbf{I}_{i,t}(\mathbf{u})\) provides the corresponding point color,
\(\Omega_i\subset\mathbb{R}^2\) denotes the pixel domain of the \(i\)-th input image,
\(M_{i,t}(\mathbf{u})\in\{0,1\}\) is the valid-depth mask, and
\(\mathcal{F}\) denotes the point-cloud filtering and fusion operation.

We use the resulting frame-wise point clouds to initialize the Gaussian representation, which is then optimized on the four observed views using the original STGS reconstruction objective~\cite{li2024spacetimegaussianfeaturesplatting}, producing an initial dynamic Gaussian field $\mathcal{G}$.



\subsection{Video-Diffusion-Guided Iterative Refinement}\label{subsec:video-diffusion}
\label{sec:diffusion}
Although dense initialization improves geometric coverage in observed regions, areas that are weakly observed or entirely absent from the four training views may still exhibit artifacts such as blur and holes. We therefore introduce a second refinement stage that leverages a pretrained video restoration model to generate view-consistent pseudo-observations along novel camera trajectories. 

\paragraph{Reference-guided video restoration}
Given the initially reconstructed 4DGS representation $\mathcal{G}$,
we introduce a pretrained CogVideoV2V~\cite{yin2025gsfixerimproving3dgaussian} model for iterative refinement. 
For each timestamp $t$ and camera pair $(i,j)$, we render a novel-view sequence
$\mathbf{V}_{t,ij}^{\mathrm{ren}}$ along an interpolated camera trajectory.
The rendered sequence, together with the two endpoint observations
$\mathbf{I}_{t}^{i}$ and $\mathbf{I}_{t}^{j}$, is fed into the frozen video
restoration model:
%
$
    \widetilde{\mathbf{V}}_{t,ij}
    =
    \mathcal{F}_{\phi}
    \left(
        \mathbf{V}_{t,ij}^{\mathrm{ren}},
        \mathbf{I}_{t}^{i},
        \mathbf{I}_{t}^{j}
    \right),
$
where $\widetilde{\mathbf{V}}_{t,ij}$ is cached as pseudo-supervision for
refining the corresponding 4DGS scene. 

\paragraph{Novel-view trajectory sampling.}
We construct six pairwise trajectories between the four input cameras. For a camera pair $(i,j)$, we interpolate the camera centers and rotations using linear interpolation and spherical linear interpolation(SLERP), respectively:
\begin{equation}
\begin{aligned}
    \mathbf{c}_{ij}(\alpha)
    &=
    (1-\alpha)\mathbf{c}_{i}
    +
    \alpha\mathbf{c}_{j}
    +
    A\sin(\pi\alpha)\mathbf{n}_{ij}, \\
    \mathbf{q}_{ij}(\alpha)
    &=
    \operatorname{SLERP}
    \left(
        \mathbf{q}_{i},
        \mathbf{q}_{j};
        \alpha
    \right),
    \qquad \alpha\in[0,1],
\end{aligned}
\label{eq:camera_trajectory}
\end{equation}
where $\mathbf{n}_{ij}$ denotes a direction perpendicular to the camera
baseline, and $A$ is a small randomly sampled perturbation amplitude.
We set $A=0$ for the six straight-line trajectories and additionally
apply $A>0$ to the two diagonal cameras to improve spatial coverage.




\paragraph{Cached pseudo-supervision}
Invoking CogVideoV2V at every optimization iteration would be computationally expensive. Moreover, its generation stochasticity could introduce inconsistent pseudo-targets across iterations. We therefore cache the restored videos and reuse them for multiple optimization steps. As the 4D Gaussian representation improves, we refresh the cache every $K$ iterations so that the pseudo-targets remain aligned with the progressively refined renderings while reducing the cost of repeated video restoration.


\begin{figure*}[t]
\centering
\includegraphics[width=\linewidth]{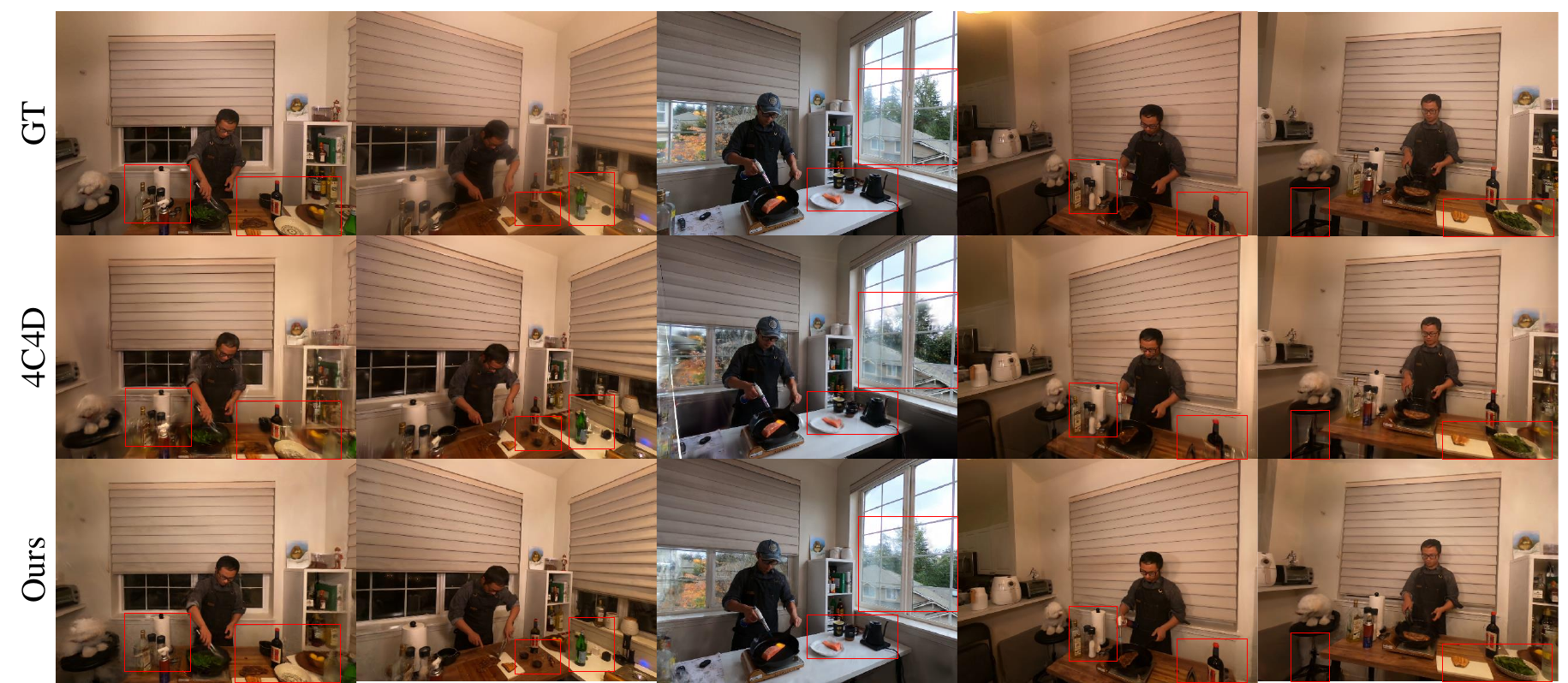}
\caption{\textbf{Qualitative visualization comparisons on the Neural3DV dataset.}}
\label{fig:fig4}
\end{figure*}

For a mini-batch $\mathcal{B}_{\mathrm{gen}}$ sampled from the cache, we define the generative supervision loss and the final objective as
\begin{equation}
\begin{aligned}
    \mathcal{L}_{\mathrm{gen}}
    &=
    \frac{1}{\lvert\mathcal{B}_{\mathrm{gen}}\rvert}
    \sum_{(t,n)\in\mathcal{B}_{\mathrm{gen}}}
    \left\|
        \mathcal{R}
        \bigl(\mathcal{G};t,\boldsymbol{\pi}_{t,n}\bigr)
        -
        \widetilde{\mathbf{I}}_{t,n}
    \right\|_{1}, \\
    \mathcal{L}_{\mathrm{total}}
    &=
    \mathcal{L}_{\mathrm{STG}}
    +
    \lambda_{\mathrm{gen}}\mathcal{L}_{\mathrm{gen}}.
\end{aligned}
\label{eq:refinement_objective}
\end{equation}
where $t$ and $n$ index the timestamp and novel view, respectively.
$\mathcal{R}(\mathcal{G};t,\boldsymbol{\pi}_{t,n})$ denotes the image
rendered from $\mathcal{G}$ at pose $\boldsymbol{\pi}_{t,n}$, while
$\widetilde{\mathbf{I}}_{t,n}$ is its CogVideoV2V-generated pseudo-target.
$\mathcal{L}_{\mathrm{STG}}$ denotes the original reconstruction loss, and
$\lambda_{\mathrm{gen}}=0.5$ balances the generative supervision. 
During refinement, the restored trajectory videos are cached and
periodically refreshed using the updated 4DGS representation.

\section{Results}
\label{sec:result}



\paragraph{Dataset and Baselines.}
We evaluated \titlePrefix{} on Neural3DV, which contains \textit{six} dynamic scenes captured by 18--21 cameras at (2704 $\times$ 2028) resolution and 30 FPS. 
Following the 4C4D evaluation protocol, we use the four most spatially separated cameras (IDs 1, 10, 13, and 20) as training views, while the remaining views are reserved for testing.
We compare against STGS, 4DGS, 4DGaussians, and 4C4D using results reported under the same protocol.
\paragraph{Qualitative and quantitative results.}
As shown in Fig.~\ref{fig:fig4}, our method preserves sharper details while reducing artifacts and blur. The quantitative results in Table~\ref{tab:neural3dv_comparison} further demonstrate that \titlePrefix{} outperforms the current state-of-the-art method, 4C4D, and substantially improves upon its STGS backbone.

\begin{table}[t]
    \centering
    \caption{Quantitative comparison on the Neural3DV dataset.}
    \label{tab:neural3dv_comparison}
    \setlength{\tabcolsep}{6pt}
    \renewcommand{\arraystretch}{1.05}

    \resizebox{\linewidth}{!}{
    \begin{tabular}{lcccc}
        \toprule
        Method
        & PSNR $\uparrow$
        & DSSIM$_1$ $\downarrow$
        & DSSIM$_2$ $\downarrow$
        & LPIPS $\downarrow$ \\
        \midrule

        \rowcolor{baselinecolor}
        STGS\cite{li2024spacetimegaussianfeaturesplatting}
        & 17.70
        & 0.158
        & 0.107
        & 0.325 \\

        4DGS\cite{yang2024realtimephotorealisticdynamicscene}
        & 20.60
        & 0.143
        & 0.094
        & 0.244 \\

        4DGaussians\cite{wu20244dgaussiansplattingrealtime}
        & 20.82
        & 0.117
        & 0.077
        & 0.190 \\

        4C4D\cite{zhou20264c4d4camera4d}
        & 22.29
        & 0.098
        & 0.062
        & 0.146 \\

        \midrule

        \rowcolor{ourscolor}
        \titlePrefix{} (Ours)
        & \textbf{24.18}
        & \textbf{0.086}
        & \textbf{0.053}
        & \textbf{0.128} \\

        \bottomrule
    \end{tabular}
    }
\end{table}

\paragraph{Ablation study.}
We analyze the contributions of dense initialization and diffusion-guided
refinement in Table~\ref{tab:ablation}. Replacing per-frame COLMAP
reconstruction with MVSAnywhere-based dense initialization improves PSNR by almost 4dB
while reducing point-cloud preprocessing time from
approximately 20 to 3 minutes. We further evaluate the restored
CogVideoV2V outputs directly against held-out ground-truth views. 
The restoration model improves both the initial and the refined 4DGS renderings. Distilling the first-round restored pseudo-targets improves the 4D representation to 24.18 dB, while a second restoration round further enhances the refined renderings, demonstrating a positive iterative refinement effect.

\section{Conclusions and Future Work}
\label{sec:conclusions}

We presented \titlePrefix{}, a generative 4D Gaussian splatting framework for
dynamic scene reconstruction from only four views. By combining dense
point-cloud initialization with iterative video-based refinement, \titlePrefix{}
improves reconstruction completeness and visual quality in sparsely observed
regions. 

The refinement stage remains computationally expensive and may be affected
by imperfect generation. Future work will explore more efficient and
reliable supervision on diverse real-world scenes.

\begin{acks}
This project is supported by the MOE Translational and Innovation Fund (TIF) (MOE2023\_TIF-0006) and partially supported by the MOE Academic Research Fund Tier 1 of Singapore. 
\end{acks}


\begin{table}[t]
\centering
\caption{Stage-wise ablation on Neural3DV. The restored
pseudo-target is evaluated directly at held-out test views.
}
\label{tab:ablation}
\setlength{\tabcolsep}{3.5pt}
\resizebox{\columnwidth}{!}{%
\begin{tabular}{lccccc}
\toprule
Method
& PSNR $\uparrow$
& DSSIM$_1$ $\downarrow$
& DSSIM$_2$ $\downarrow$
& LPIPS $\downarrow$
& Prep. Time $\downarrow$ \\
\midrule
STGS w/ COLMAP
& 17.702 & 0.158 & 0.107 & 0.325 & $\sim$20 min \\
STGS w/ MVS
& 21.674 & 0.109 & 0.071 & 0.165 & $\sim$3 min \\
CogVideoV2V restoration (init. 4DGS)
& 23.563 & 0.105 & 0.071 & 0.107 & -- \\
Full
& 24.180 & 0.086 & 0.053 & 0.128 & -- \\
CogVideoV2V restoration (refined 4DGS)
& 24.334 & 0.096 & 0.065 & 0.099 & -- \\
\bottomrule
\end{tabular}
}
\end{table}

\bibliographystyle{ACM-Reference-Format}
\bibliography{sample-bibliography}

\appendix

\end{document}